# CyclingNet: Detecting cycling near misses from video streams in complex urban scenes with deep learning

Mohamed R. Ibrahim[1], James Haworth, Nicola Christie and Tao Cheng
Department of Civil, Environmental and Geomatic Engineering, University College London (UCL)
[1] Corresponding author: mohamed.ibrahim.17@ucl.ac.uk

*Abstract*— Cycling is a promising sustainable mode for commuting and leisure in cities, however, the fear of getting hit or fall reduces its wide expansion as a commuting mode. In this paper, we introduce a novel method called CyclingNet for detecting cycling near misses from video streams generated by a mounted frontal camera on a bike regardless of the camera position, the conditions of the built, the visual conditions and without any restrictions on the riding behaviour. CyclingNet is a deep computer vision model based on convolutional structure embedded with self-attention bidirectional long-short term memory (LSTM) blocks that aim to understand near misses from both sequential images of scenes and their optical flows. The model is trained on scenes of both safe rides and near misses. After 42 hours of training on a single GPU, the model shows high accuracy on the training, testing and validation sets. The model is intended to be used for generating information that can draw significant conclusions regarding cycling behaviour in cities and elsewhere, which could help planners and policy-makers to better understand the requirement of safety measures when designing infrastructure or drawing policies. As for future work, the model can be pipelined with other state-of-the-art classifiers and object detectors simultaneously to understand the causality of near misses based on factors related to interactions of road-users, the built and the natural environments.

## 1. INTRODUCTION

Cycling for commuting or leisure is a growing transport mode across the globe. Its benefit for health and the natural environment have driven different policies to promote it and build more infrastructure for cycling in cities (de Hartog et al., 2010; Juhra et al., 2012). However, the modal share of cycling remains low in comparison to other transport modes in part because it is perceived as a dangerous activity, regardless of its benefits (Blaizot et al., 2013). It has been found that in the UK, for instance, this fear of falling or being in a collision with other road-users limits wider expansion of cycling as a transport mode (Aldred, 2016; De Rome et al., 2014; Winters and Branion-Calles, 2017). However, in many countries, quantitative analysis of cycling safety is difficult because the low mode share of cycling results in few recorded incidents. To address this data gap, scholars have analysed the occurrence of near-misses as a proxy for incidents due to their higher frequency, which some studies estimate is as high as 0.172 incidents per mile (Aldred, 2016).

Quantitative data on near misses is usually collected in one of three ways; self-reported surveys or questionnaires, site observation (e.g. at an intersection) and naturalistic studies (Ibrahim et al., 2020b). Of these, naturalistic studies can provide the richest data through bike-mounted sensors such as video cameras, GPS, range sensors and accelerometers. This type of data is also routinely collected by many cyclists, who use action cameras for safety in the same way car drivers use dashcams, reporting incidents to the police. However, analysis of these data, particularly video scenes, is usually done manually, which is a labour-intensive process that limits broader applicability of the method. Accordingly, finding a method that could automatically detect near misses and their risk factors from naturalistic cycling data would transform its applicability from small scale studies to mass applications to crowdsourced video streams and real-time operation. Such a system would be of significant interest for bike riders and planners and policy-makers alike.

The field of Artificial intelligence (AI), specifically, the domain of deep learning and computer vision, has the potential to address this gap. Various models have been developed to recognise a wide spectrum of human actions, activities, or body poses in complex settings from untrimmed video streams. *Near misses result from unsafe interaction between a number of road-users or obstacles that caused a risky situation for the person on a bike and subsequently an instant action or a group of actions (e.g. swerving, stopping, turning left or right, etc.) needed to be taken to avoid a crash.* Therefore, a computer vision system that can detect this type of interaction will need to not only detect actions but understand how the actions of individuals within a scene interact to produce risk.

In this article, we introduce a new method called CyclingNet to detect cycling near misses from untrimmed video streams in complex urban scenes. The model is trained on video streams of a frontal camera on a moving bicycle, either mounted in the helmet of the cyclist or the bicycle handlebar without any restrictions towards the camera angle, the physical or the visual conditions of the built environment. Our goal is to provide a fast algorithm that can be deployed on a camera to be used in real-time or near-real-time settings for detecting and evaluating near misses in different outdoor scenes. Our main contributions are:

- Automating the detection of cycling near misses in a near real-time detection.
- A novel end-to-end deep model for recognising cycling near misses from untrimmed video streams in complex urban settings.
- A human-labelled large scale dataset for classifying video streams of moving bicycles -at a frame level- of near miss and non-near miss.
- A comprehensive set of experiments to evaluate the different architectures of deep models that can be used as a baseline for future research in this study domain.

At this point, it should be noted that the detection of risk factors associated with near miss events is not the topic of this paper, but the task can also be accomplished using computer vision algorithms (see, e.g. (Mohamed R Ibrahim et al., 2019; Mohamed R. Ibrahim et al., 2019)).

After the introduction, the paper is structured in six sections. The second section reviews relevant work of the current near misses methodologies and the advances of computer vision in action recognition. In section 3, we explain our method and the materials used. In section 4, we show our model results, baseline analysis, and the evaluation metrics. Afterwards, in section 5, we discuss our results with the current literature. We also highlight the state-of-the-art of CyclingNet and its limitations. Last, in section 6, we conclude and we give our remarks for future work.

## 2. RELATED WORK

### 2.1 Limits of the current methods for analysing cycling near misses

The term cycling near miss is subjective and an individual's perception of an event as a near miss may differ based on their experience level, personal characteristics, and their perception of risk. 'Near collision' (Johnson et al., 2010), 'perceived crash risk' (Chaurand and Delhomme, 2013; Strauss et al., 2013), 'perceived traffic risk' (Sanders, 2015), or 'near miss' (Aldred, 2016; Poulos et al., 2012) are all terms often used to describe and address near misses in literature. In this research, we use the definition from (Ibrahim, Haworth, Christie, et al., 2020, p.4), which describes a near miss as *"a situation in which a person on a bike was required to act to avoid a crash, such as braking, speeding, swerving or stopping. In some cases, the definition may be extended to include those events that caused the person on the bike to feel unstable or unsafe, such as a close pass or tailgating"*. This broader definition is used because it encompasses the range of near-miss events that have been identified in the literature, including (1) close pass, (2) a near left or right hook, (3) someone pulling in or out, (4) a near-dooring, (5) swerve around an obstruction, (6) pedestrian steps out, (7) someone approaching head-on, or (8) tailgating (Aldred and Goodman, 2018a), while also allowing for other unforeseen events that involve action to avoid a crash.

### 2.1 Current methods for analysing cycling near misses

In the literature, near misses have been analysed using different types of observational studies, which can be categorised according to (Ibrahim et al., 2020b) as 1) self-report studies using surveys or questionnaires (Aldred, 2016; Aldred and Goodman, 2018b; Paschalidis et al., 2016); 2) video analyses at specific sites such as intersections (Lehtonen et al., 2016; Vansteenkiste et al., 2016), and 3) naturalistic studies where video stream data is collected as people cycle (Dozza and Werneke, 2014a; Johnson et al., 2010, 2013). In general, the naturalistic approach has shown the most progress in analysing road conflicts, near misses, and crashes due to the nature of the data collected (Dozza et al., 2016; Dozza and Werneke, 2014b; Schleinitz et al., 2017). In this approach, a group of participants carries out their daily activities using bikes instrumented with cameras and sensors. Rich data related to the environment, riding behaviour and interaction with other road users can be collected. However, due to the need to manually label video data, current naturalistic studies are labour intensive and are limited in both transferability from one location to another, and scaling up to cover a wider region or larger number of participants. Thus, it is difficult to draw objective conclusions that can allow either a change in cities' policies or road-users' behaviours to provide a safer and more inclusive environment.

### 2.2 Action recognition from video streams

While the issue of detecting cycling near misses from moving bicycles in real-world settings has not been addressed in the literature, there is a well-established body of knowledge on action recognition from video streams. Action recognition using CV typically involves two steps: 1) extracting and encoding features and 2) classifying features into action classes (Kang & Wildes, 2016). In recent years, Convolutional Neural Networks (CNNs) have been applied to the task with great success. Different strategies have been adopted in designing the models' architecture to extract features that could enhance the training and inference of the model. Some models, for instance, rely on spatial features to classify actions, whereas others include both temporal and spatial aspects of the scene to classify and localise multiple actions. State-of-the-art models are tested on benchmark datasets, with UCF-101 and HMDB-51 being popular for action recognition tasks. UCF-101 is a dataset of 13,320 YouTube videos broken down into 101 action categories and a further 25 groups. HMDB-51 is similar, with 51 actions in 6,849 clips. For example, Simonyan & Zisserman (2014) introduced an action recognition model relying on a two-stream convolution structure, exploiting both RGB data and optical flow[1]. The model is evaluated on UCF-101 and HMDB-51, achieving a top-performance of 87.9% on UCF-101 dataset. Wang et al. (2015) introduced deep convolutional descriptors based on trajectory pooling. The model is structured and trained as a two-streams convolutional structure in which features are extracted based on the RGB image and the tracked trajectories in the sequential frames. The model achieved the best accuracies of 65.9% and 91.5% on UCF-101 and HMDB-51datasets respectively. Ng et al. (2015) introduced a hybrid model of the convolutional structure and long-short term memory (LSTM) blocks to classify actions based on their temporal structure, achieving a result of 88.6% and 82.6% on UCF-101 dataset, with and without optical flow data, respectively. Most significantly, Wu et al. (2015) introduced a spatio-temporal model relying on the two-streams network, utilising both RGB frames and optical flows. The model architecture is based on integrating LSTM on top of the convolution structures for the two streams. The model achieves a top score of 91.3% on the UCF-101 dataset.

Recently, approaches have been also introduced to tackle actions in video streams besides the 2D convolution structure and LSTM units. For example, relying on 3D convolutional structure (where time is the third dimension), Diba et al. (2017) introduced a new temporal 3D CNN model relying on a

---

[1] Optical flow refers to the percieved motion of objects in a given scene in respect to relative motion of the observer and the objects in the scene (Gibson, 1950).

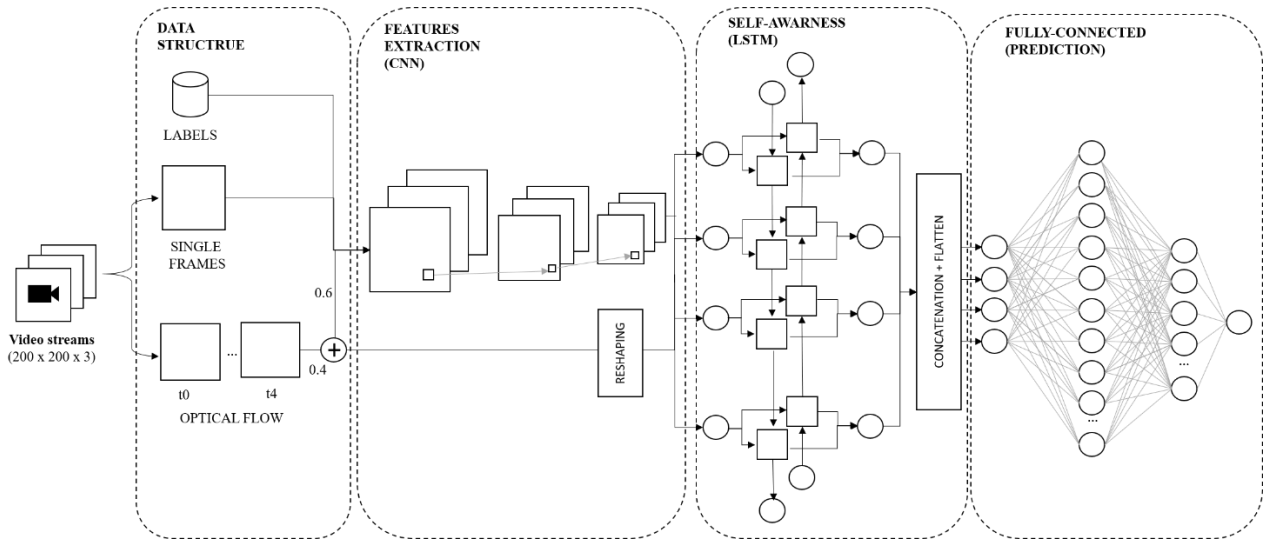

FIGURE 1
THE ARCHITECTURE FOR THE PROPOSED CYCLINGNET

Temporal Transition Layer to recognise human actions in video streams. This spatio-temporal model aimed to capture the variations of the dynamics for video representation. The model achieved top-score of 93.2% and 63.5% in UCF-101 and HMDB-51datasets respectively. Girdhar & Ramanan (2017) introduced a new attentional pooling structure that has been improved the accuracy of action recognition on various benchmarked datasets without any cost on the computation intensity or time of inference. For instance, the model achieved better performance on MPII dataset than the previous methods with 12.5% relative overall improvement (Andriluka et al., 2014).

There are variations of action recognition models which focus mainly on understanding human activities rather than the overall perception of the interaction between different agents or the clue of the scene in the case of the stated issue of near misses (Kang & Wildes, 2016). In summary, not only the architecture of action recognition models vary, but also the training process and the data fusion approach. Some models have been trained in an end-to-end network, whereas others are designed and trained in a two-stream network with an early or late-stage fusion of data types (RGB frames, optical flow data, etc.). While complexities of these model have yielded better accuracies in a given task, specifically, relying on the two-streams network, however, these differences have consequences on the trade-off between model accuracy, complexity, and time needed for inference which determine whether the model could function in real-time. Video recognition, however, remains a challenging task due to the high variance between the sequential images and inter-classes and the low-resolution of videos (Wang et al., 2015b).

3. METHODS

*3.1 Model requirements*

As stated in section 2.1, we define a near miss as *"a situation in which a person on a bike was required to act to avoid a crash, such as braking, speeding, swerving or stopping. In some cases, the definition may be extended to include those events that caused the person on the bike to feel unstable or unsafe, such as a close pass or tailgating"* (Ibrahim, Haworth, Christie, et al., 2020, p.4). In order to identify these events, a Computer vision algorithm must be capable of distinguishing such a set of instant actions from normal riding behaviour, which may also include actions similar to those taken during a near miss.

Near misses can be seen as instant actions that take place by other objects in the scene. Accordingly, there are three main elements that the model needs to learn in order to recognise near misses: *1) The relative motions of the elements in the scene, 2) the spatial structure of the scene, and 3) memory to recognise what happened in the past.*

Subjectively, understanding the change in motion could lead to a better way of understanding the actions related to both safe and unsafe rides since each object conserves its motion between consecutive frames and neighbouring pixels are more likely to conserve similar motion. Accordingly, combining street-level frame images with their optical flow for a number of consecutive frames may lead to a better approach to recognise near misses from video streams.

*3.2 Model architecture*

In order to respond to the aforementioned requirements, we propose the CyclingNet model. The CyclingNet is a novel single-stream spatio-temporal deep model that is trained in an end-to-end fashion. It aims to include the features of two-streams networks by including the spatio and temporal aspects of the video stream, whereas providing an inference in near real-time similar to the single-stream networks. Its algorithms comprise four main sections; data structure, features extraction, self-awareness, and integration and prediction. Figure 1 shows the order of the main algorithms and how the model is structured.

*3.2.1 Data structure*

As input, the model takes video streams typically produced by cameras mounted on a moving bike, which may have varying angles, field of views, rider speeds and filtering processes (e.g for stabilisation or extraction of a region of interest). The video streams are resized into 240 X 320 X 3 tensors of single-frame images, in addition to a computed dense optical flow for each pixel in two consecutive frames.

For a given pixel $P_{(x,y,t)}$ that moves a (d) distance of $(dx, dy)$, the change in P, assuming that P does not change its intensity, can be calculated as:

$$P_{(x,y,t)} = p(x + dx, y + dy, t + dt) \tag{1}$$

By dividing the right side with *dt* and using Taylor approximation technique, we estimate the optical flow as:

$$f_x u + f_y v + f_t = 0 \tag{2}$$

given that: $f_x = \frac{\partial f}{\partial x}, f_y = \frac{\partial f}{\partial y}, u = \frac{dx}{dt}, v = \frac{dy}{dt}$

Where $(f_x)$ and $(f_y)$ are the image gradients, $(f_t)$ is the gradient over time, whereas $(u)$ and $(v)$ are unknown.

In order to solve this equation, with several unknown gradients, we used Gunner Farneback's algorithm (Farnebäck, 2003), in which he approximates each neighbourhood by a quadratic polynomial. Consequently, a new signal can be constructed based on a global displacement, which can be computed based on equating the coefficients of the yields of the quadratic polynomials.

The outputted optical flow vectors $(u, v)$ are an array of two channels, which can be visualised in a colour image, with a magnitude as the value plane, and direction as the hue value.

After computing the optical flow $(f_{o\prime(t)})$ for a given time $(t)$, the data of the RGB images $(f_{rgb})$ are truncated for each video file to start with the 4$^{th}$ frame in the frame sequence and wrapped with the four timestamps of the frames of optical flows[$t_i$, $t_{i-1}$, $t_{i-2}$, $t_{i-3}$] in a proportion of 0.5 to 0.5 respectively. The input $(x_{(t0)})$ is defined as:

$$x_{(t_i)} = \frac{f_{rgb\prime(t_i)}}{2} + \frac{f_{o\prime(t_i)} + f_{o\prime(t_{i-1})} + f_{o\prime(t_{i-2})} + f_{o\prime(t_{i-3})}}{8} \tag{3}$$

There are two reasons for selecting and optimising these hyperparameters: Firstly, to add the time dimension to the spatial structure of each street-level image, and secondly, to control and reduce the information and the number of features and textures that are not useful for detecting near misses (i.e the textures of people, cars, building, etc.). We experimented with the values of the combined ratio, based on trial and error to optimise the overall fitness and performance of the model when detecting near misses.

The output data is structured and reshaped in four-dimensional tensors (timestamps, width, height, channels), in addition to embedding the four optical flow steps with the single-frame images. Such an approach means the dimension of time can be utilised and seen either in the spatial structure of the image (fusion with four previous steps of the optical flows) or the series of the data (the length of timestamps). Both approaches will be utilised and discussed thoroughly in the algorithms of CyclingNet in the two upcoming sections.

*3.2.2 Extracting features*

The goal of this part of the model is to extract mainly spatial features from the single frame images, bearing in mind the fused data of the optical flows of the previous four steps. The architecture of this section comprises three consecutive blocks of convolutional structure, each having different sets of structure and hyperparameters and initialised by "He normal" initialisation technique to provide more efficient and faster gradient descent (He et al., 2015). Generally, the choices of the presented hyperparameters are made based on trials and errors, and the most common practice for training Convolutional models. Nevertheless, different models with different hyperparameters will be trained and presented as base models for further evaluating the introduced methods, in the results section (Section 4.2).

Block one consists of two 2D convolution layers of a kernel size (24 X 5 X 5), (36 X 5 X 5) respectively, and a subsampling size of (2 X 2). They are activated based on a Rectified Linear Unit (ReLU). These two CNN layers are followed by a 2D Max-Pooling layer of pool size of (2 X 2) and a Batch-normalization layers of the momentum of 0.99 and epsilon of 0.001. It is feed with single frame images with the embedded optical flow steps.

Similar to block one, block two consists of two 2D Convolution layers, however, a kernel size (48 X 5 X 5), (64 X 3 X 3) respectively, and a subsampling size of (2 X 2). They are activated based on a Rectified Linear Unit (ReLU). They are also followed by a 2D Max-Pooling layer of pool size of (2 X 2) and a batch-normalization layers of the momentum of 0.99 and epsilon of 0.001.

Block three consists of a single convolution layer of a kernel size (128 X 3 X 3) subsampled with (2 X 2) and activated by a ReLU function. It also followed by a 2D Max-Pooling layer of pool size of (2 X 2) and a Batch-normalization layers of the momentum of 0.99 and epsilon of 0.001.

*3.2.3 Spatial and temporal awareness*

If the algorithms of detecting near misses rely only on the features of the previous section (the convolution structure), based on experiments, the results will be sensitive to the changes of the spatial structure at the local context. In other words, the model would not have taken into account the global context of the inputted features that ensure stability and accuracy for training and inference. For this reason, designing the architecture of CyclingNet further to be aware of both local and global spatial and temporal structure is important.

This part of the model comprises one bidirectional Long-Short Term Memory (LSTM) block, followed by a regulated self-attention layer. The LSTM block consists of 128 units, and a dropout regulation of a size of 0.3 to avoid over the fitness of the model. However, the goal is not only considering the sequence of the defined timestamps but also considering the context for each timestamp, therefore, a self-attention mechanism is essential to ensure the balance for both global and local context when describing a given scene.

Generally, a unidirectional LSTM has shown great progress in extracting features related to sequential data to predict future states (Goodfellow et al., 2017; LeCun et al., 2015). Unlike a traditional recurrent layer, LSTM can learn long-term dependencies without suffering from issues related to vanishing gradient. This internal recurrence, so-called self-loop enabled the previous vectors to create paths, in which the gradient can move forward for a long duration without vanishing issues. Nevertheless, most recently, it has also been shown to improve the overall performance of the model when predicting even a given state without timestamps by learning not only the spatial structure of a given vector but also the short-term dependences among the input given vector as the time constants are output by the LSTM itself. Accordingly, this allows the time scale to change based on the input sequence, even if the LSTM units have fixed parameters.

TABLE 1
CYCLINGNET LAYER STRUCTURE AND HYPERPARAMETERS

| Block | Layer | Output shape | Number of parameters[2] |
|---|---|---|---|
| Input | $x_{(t_i)}$ | (None[1], 240, 320, 3) | 0 |
| Block 1 | conv2d_1 (Conv2D) | (None, 118, 158, 24) | 1824 |
| | conv2d_2 (Conv2D) | (None, 57, 77, 36) | 21636 |
| | max_pooling2d_1 (MaxPooling2) | (None, 28, 38, 36) | 0 |
| | batch_normalization_1 (BatchNorm) | (None, 28, 38, 36) | 144 |
| Block 2 | conv2d_3 (Conv2D) | (None, 12, 17, 48) | 43248 |
| | conv2d_4 (Conv2D) | (None, 10, 15, 64) | 27712 |
| | max_pooling2d_2 (MaxPooling2) | (None, 5, 7, 64) | 0 |
| | batch_normalization_2 (BatchNorm) | (None, 5, 7, 64) | 256 |
| Block 3 | conv2d_5 (Conv2D) | (None, 3, 5, 128) | 73856 |
| | max_pooling2d_3 (MaxPooling2) | (None, 1, 2, 128) | 0 |
| | batch_normalization_3 (BatchNorm) | (None, 1, 2, 128) | 512 |
| Reshape | | (None, 2, 128) | 0 |
| Block 4 | Attention (SeqSelfAttention-LSTM) | (None, 2, 1024) | 1048577 |
| | bidirectional_1 (Bidirection-LSTM) | (None, 2, 1024) | 2625536 |
| | dropout_1 (Dropout) | (None, 2, 1024) | 0 |
| Flatten | | (None, 2048) | 0 |
| Block 5 | dense_1 (Dense) | (None, 256) | 524544 |
| | dropout_2 (Dropout) | (None, 256) | 0 |
| | dense_2 (Dense) | (None, 64) | 16448 |
| | dropout_3 (Dropout) | (None, 64) | 0 |
| Output | dense_3 (Dense) | (None, 1) | 65 |

[1]None values represent the total number of samples in 64 batches in training and validation sets.
[2]The total trainable parameters: 4,383,902 and the non-trainable params: 456

To extract long-term dependences, the self-loops of the LSTM units can be controlled by three gated units: 1) forget gate ($f_i^{(t)}$), external input gate ($g_i^{(t)}$), and an output gate ($q_i^{(t)}$). First, ($f_i^{(t)}$) can be explained for a given cell (i) and time (t), whereas it is fitted to a scaled value in the interval [0,1] and a sigmoid activation unit ($\sigma$) as:

$$f_i^{(t)} = \sigma\big(b_i^f + \sum_j U_{i,j}^f x_j^{(t)} + \sum_j W_{i,j}^f h_j^{(t-1)}\big) \quad (4)$$

given that $h^{(t)}$ represents a vector that contains the outputs of all the LSTM cells for the current hidden layer, $x^{(t)}$ represents the current input vector, $W^f$ represents the recurrent weights for the forget gates, $U^f$ represents the input weights and last, $b^f$ represents the biases of the forget gates.

Second, to update the LSTM internal state, a conditioned weight of the self-loop ($f_i^{(t)}$) is computed as:

$$s_i^{(t)} = f_i^{(t)} s_i^{(t-1)} + g_i^{(t)} \sigma\big(b_i + \sum_j U_{i,j} x_j^{(t)} + \sum_j W_{i,j} h_j^{(t-1)}\big) \quad (5)$$

given that U is the input weights, b is the bias vector, W represents the current weights into the LSTM cell. Similar, to ($f_i^{(t)}$), the external input gate ($g_i^{(t)}$) is computed, however with it is a parameter:

$$g_i^{(t)} = \sigma\big(b_i^g + \sum_j U_{i,j}^g x_j^{(t)} + \sum_j W_{i,j}^g h_j^{(t-1)}\big) \quad (6)$$

Last, the output gate ($q_i^{(t)}$) is used to control and shut off the LSTM cell output ($h_i^{(t)}$) with a sigmoid unit, in which the $h_i^{(t)}$ is defined as:

$$h_i^{(t)} = \tanh(s_i^{(t)}) q_i^{(t)} \quad (7)$$
$$q_i^{(t)} = \sigma\big(b_i^o + \sum_j U_{i,j}^o x_j^{(t)} + \sum_j W_{i,j}^o h_j^{(t-1)}\big) \quad (8)$$

where $b^o$ is the model biases, $U^o$ is the input weights, $W^o$ is the current weight.

Unlike unidirectional LSTM units, a bidirectional LSTM layer allows the current hidden state to rely on two independent hidden states, one computed in a forward direction, named a forward LSTM, and the latter in the opposite direction, named a backward LSTM. This allows the retention of historical and current information simultaneously. This has a direct implication when detecting near misses, in which the predicted output for a given state is smoothed when compared to the previous ones without any post-prediction smoothing techniques.

Moreover, adding a self-attention mechanism to the bi-directional LSTM units allows the model to learn not only from the extracted features – whether spatial or temporal ones- but also to learn from the relations of the input sequences of the RGB image and optical flow ones by allowing the model to relate the positioning of each sequence and accordingly, learns the representation of its input (Goodfellow et al., 2017; Vaswani et al., 2017). Nevertheless, the model can learn which context to consider for a given scene to output the prediction (Xu et al., 2016). The context ($l_t$) can be computed as:

$$l_t = \sum_{t'} a_{t,t'} x_{t'} \quad (9)$$
given that:
$$h_{t,t'} = \tanh(x_t^T W_t + x_{t'}^T W_x + b_t) \quad (10)$$
$$e_{t,t'} = \sigma(W_a h_{t,t'} + b_a) \quad (11)$$
$$a_t = softmax(e_t) \quad (12)$$
where $(h_{t,t'})$ represents the hidden state of the previous step – in a given direction of the bidirectional LSTM- that is fitted to a simple forward neural model $(e_{t,t'})$, $(a_t)$ is the amount of the attention that the output at a given state should consider for the previous activation $(\sigma)$.

*3.3.4 Model initialization and training*

After the LSTM block, the output is flattened and fed forward to two fully-connected layers of 180 and 64 neurons respectively. Both layers are activated by a ReLU function, in which a Dropout mechanism is applied for both layers with a size of 0.3. The final output layers consist of a single neuron and activated with a sigmoid function.

The model is compiled with stochastic gradient descent, relying on "adam" optimiser, with a momentum of 0.9, and an initial learning rate of 0.001. The model is set to be trained for maximum training cycles (epochs) of 100, with an early stopping technique, monitoring the change in loss with a patience value of 20 epochs.

Table 1 summaries the different layers of the CyclingNet model. It shows the transition of their input shapes and the number of hyperparameters for each layer. Overall, the model has 4,383,902 trainable parameters.

*3.3 Evaluation metrics*

The model is penalised during training, testing and validation based on a cost function of cross-entropy of error. It is defined as:
$$E = -\sum_i^n t_i \log(y_i) \quad (13)$$
given that $t_i$ represents the target vector, $y_i$ represents the predicted vector, and n represents the binary classes.

For further assessing the model performance, we computed accuracy, precision, recall, false-positive rate, and F1-score:
$$Accuracy = (TP + TN)/(TP + TN + FP + FN) \quad (14)$$
$$Precision = TP/(TP + FP) \quad (15)$$
$$Recall = TP/(TP + FN) \quad (16)$$
$$False - positive\ rate = FP/(FP + TN) \quad (17)$$
$$F1 - score = 2 \times \frac{Precision\ X Recall}{Precision + Recall} \quad (18)$$
where $FP$ represents the predicted false-positive values, $TP$ represents the predicted true-positive values, $FN$ represents the predicted false-negative values, and $TN$ represents the predicted true-negative values.

Last, it remains a challenge to compare our results with other models due to the absence of other models for detecting near misses for a moving cyclist from a street-level. We created, however, different architecture to draw a baseline for the performance of the proposed method and to show how the different architecture and hyperparameters could yield different outcomes for a given task with the same material types.

*3.4 Materials and data pre-processing*

To the best of our knowledge, there is no benchmark data set of video streams that focus on the different types of cycling near misses that is open-sourced to conduct computer vision research. Therefore, collecting our own dataset becomes the only way to train the model to detect near misses in complex environments. We collected video clips that were made available online by people on bikes on two websites: YouTube, and Road.cc. In these clips, near misses are labelled manually in the embedded frames by the sharers. Two aspects make this data a significant one for understanding near misses: First, the variation in the perceptions of near misses as defined by the clips sharers. This could allow the model to extract features related to the common trends instead of being heavily directed or biased with a small group of participants or self-labelling. Second, the variation of equipment, camera position, context, visual, and weather conditions along with the different behaviours and riding styles in these scenes are crucial for the learning process of the model, generalisation, and deployment.

After qualitatively inspecting the quality and ground truth of the embedded information of the selected clips, we collected a dataset of 74,477 sequential frames and we computed their equivalents of optical flows frames (74,469). Of these 8,567 sequential frames belong to near miss cases (11.5% of the total sequential frames) which occur at sparse intervals. They represent 209 unique near misses of an average duration of 1.3 seconds (40.9 sequential frames). We also used an additional dataset of 12,812 sequential frames for further testing, after training and validation. This dataset comprises 81 unique near miss events.

These clips include complex urban settings of different visual and weather conditions, including, day, night, and dawn/dusk time, rain, snow, and clear weather conditions. The clips also consist of variations of near misses types of different temporal scale (the duration of near miss) and various interactions with different road-users. The clips, for instance, include near misses such as close pass, a near left or right hook, someone pulling in or out, swerve around an obstruction, pedestrian steps out, and someone approaching head-on. However, they lack clips that include near-dooring and tailgating events. Figure 2 shows a sample of the sequential frames and their corresponding optical flows.

Data augmentation for deep learning has been shown as a strong indicator for enhancing the training process and accuracy of the model (Mikolajczyk and Grochowski, 2018). Accordingly, we augmented the collected data by applying several techniques such as normalisation, scaling, and horizontal flipping.

4. RESULTS

| Table 2: Classification metrics for CyclingNet | | | | |
|---|---|---|---|---|
| **Self-attention Bi CNN-LSTM** | Precision | Recall | False-positive rate | F1-score |
| **Validation set** | 0.994 | 0.995 | 0.041 | 0.994 |
| **Test set** | 0.842 | 0.927 | 0.418 | 0.883 |

*4.1 CyclingNet evaluation*

Training the model on street-level images of both safe rides and near misses took almost 2 days (42 hours) on a single GPU (Titan V). Figures 3a-3b show the losses and accuracies of the training and testing sets respectively for the self-attention bidirectional CNN-LSTM architecture. After 35 training cycles (epochs) of 100 epochs, the model has converged and the training has stopped to avoid over-fitness after no significant change on the validation loss. In table 2, we expand further on evaluating the classification of CyclingNet. The table shows a high validation in terms of precision, recall, and an F1-score, with minimum false-positive rates. The model shows a high validation in terms of true positive of the area under the curve of 0.99 and 0.84 for validation and testing sets, respectively. However, the gap between the values of the validation and testing sets can be explained due to the variations in near miss events, or the limitations of similar events that the model can learn and extract features from for future inference.

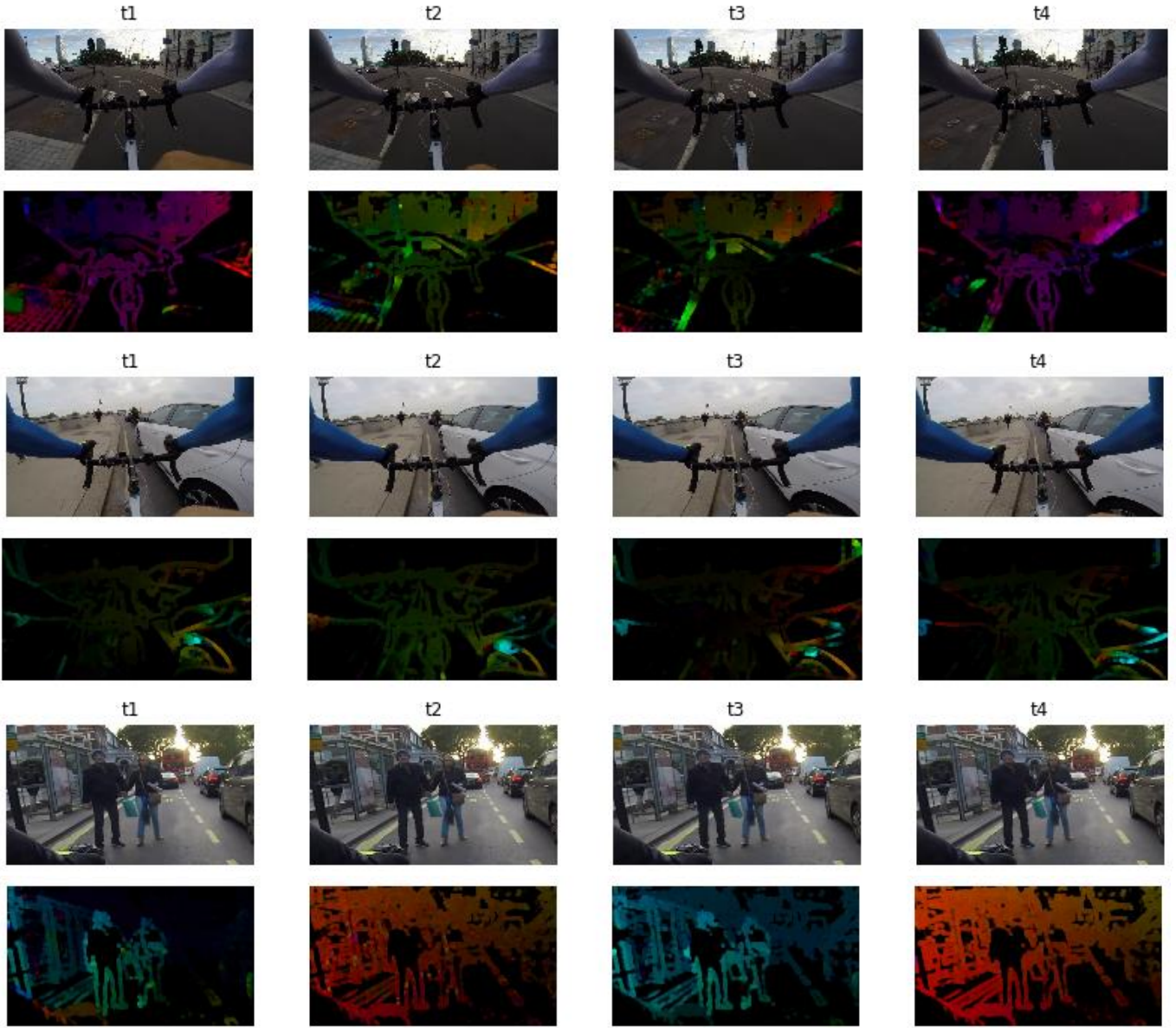

FIGURE 2
SAMPLE OF DATASET FOR THE RGB FRAMES AND THEIR OPTICAL FLOWS

### 4.2 Baseline evaluation

We experimented with adjusting optical-flow to images fusion ratio, model architecture, an optimisation technique, and post-prediction with the classification thresholds aiming to maximise temporal smoothing while reducing the global loss. We found that the global loss can be reduced even by a simple CNN architecture, however, the predicted values are prone to temporal instability. On the contrary, after applying a CNN-LSTM architecture the temporal dependences improved whereas the model outputs a smoothed prediction throughout the clip. We also found that by including bidirectional and self-attention mechanisms in the architecture of the CNN-LSTM model, the losses at the local and global levels of the training and testing datasets have improved in comparison to a CNN-LSTM model. Table 3 summarises the outcomes of the different studied architectures on the validation set, with a constant fusion ratio of 50% of the single images and optical flows.

(A)

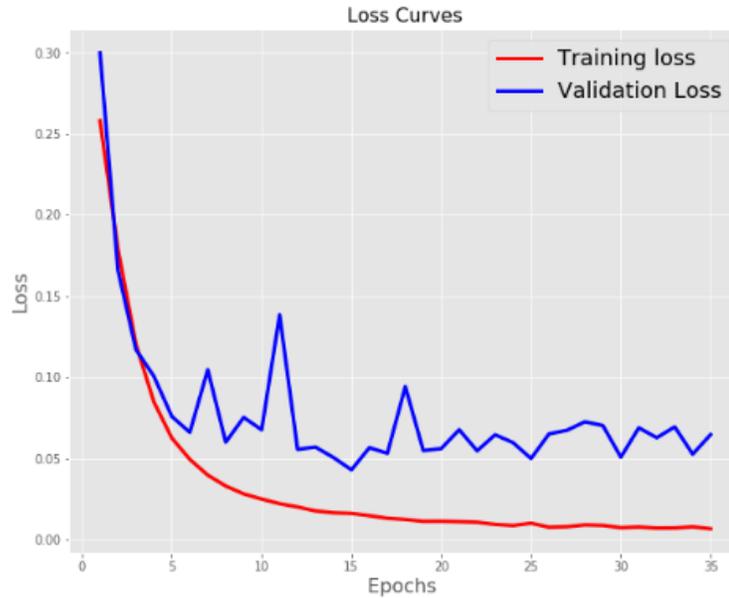

(B)

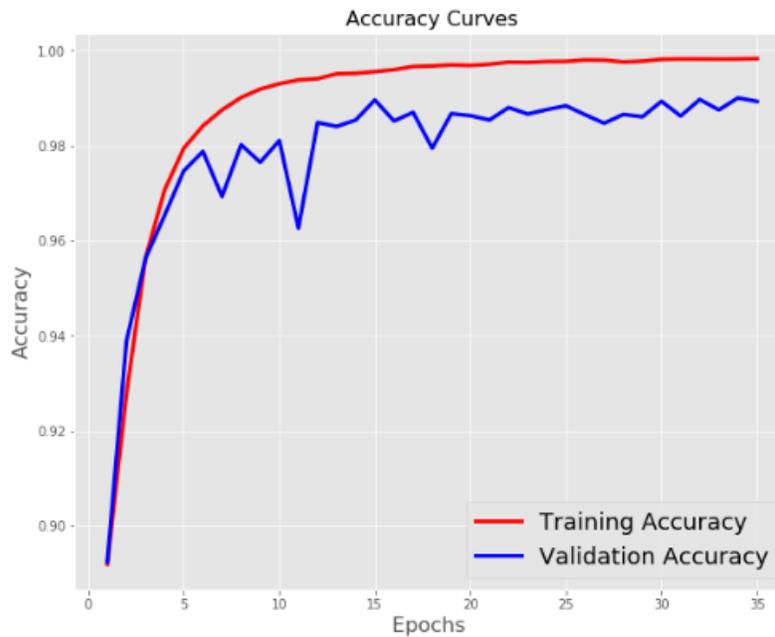

FIGURE 3
TRAINING AND EVALUATION OF CYCLINGNET

### 4.3 Scenes prediction

In figure 4, we show different clips of near misses predicted by CyclingNet. The model shows high accuracy in predicting a wide range of complex urban scenes at a different time of the day and different weather conditions. Nevertheless, the model shows high accuracy in predicting near misses including different types of near misses, such as close passes, pedestrian step in, or any risky situation with different road-users, including other people on bikes. Similarly, figure 5 shows a variation of urban scenes that has been detected as a safe ride.

| Table 3: Baseline assessment of CyclingNet | | |
|---|---|---|
| Architecture comparison | Validation Accuracy | Validation Loss |
| CNN (Block 1-3) | 86.5 % | 0.73 |
| CNN-LSTM | 97% | 0.15 |
| Self-attention CNN-LSTM | 96 % | 0.20 |
| **Self-attention Bi CNN-LSTM** | **98.9 %** | **0.06** |

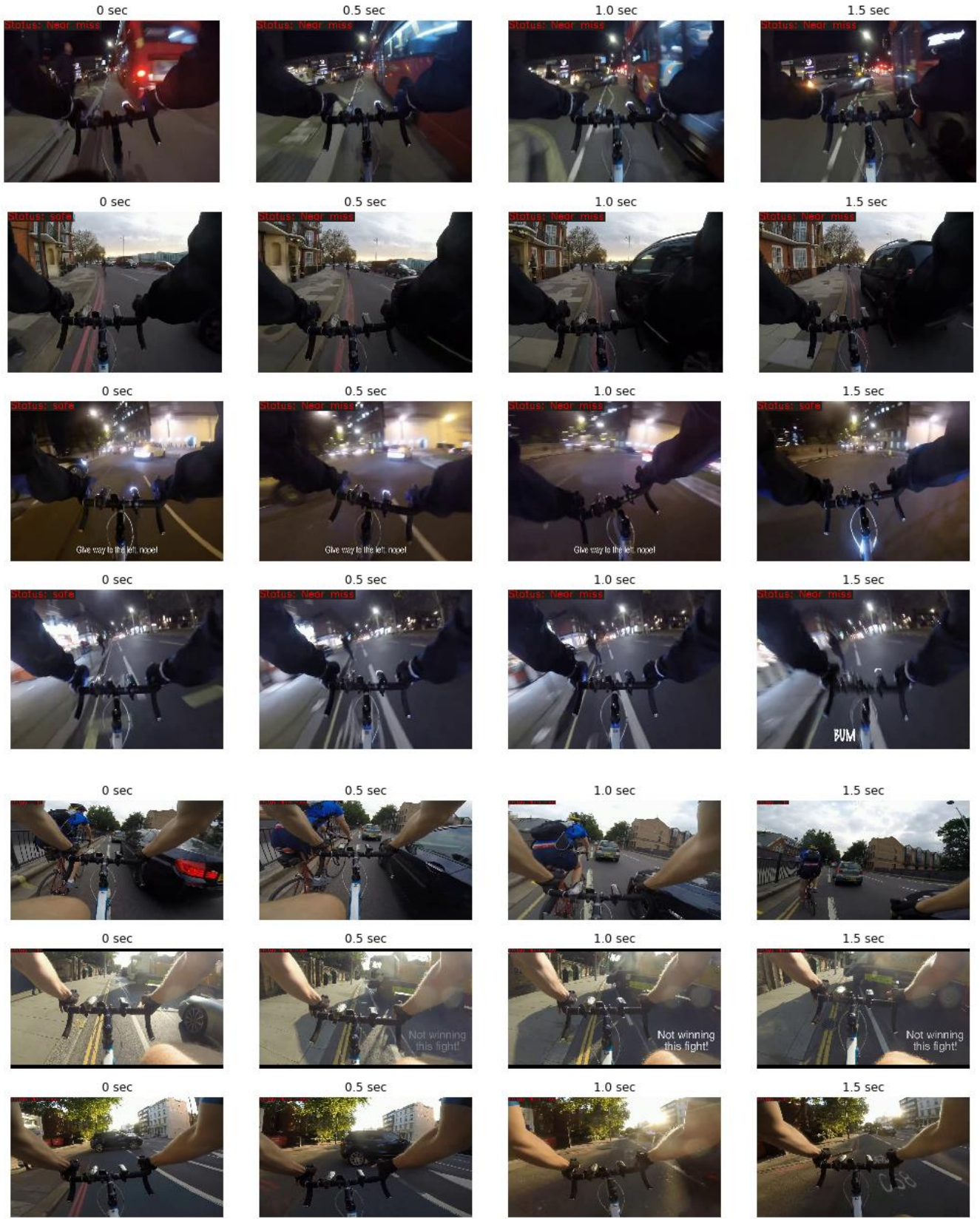

FIGURE 4
EXAMPLES OF PREDICTED CYCLING NEAR MISSES BY CYCLINGNET

SEQUENTIAL IMAGES OF SELECT FRAMES WITH A LAG OF 0.5 SEC

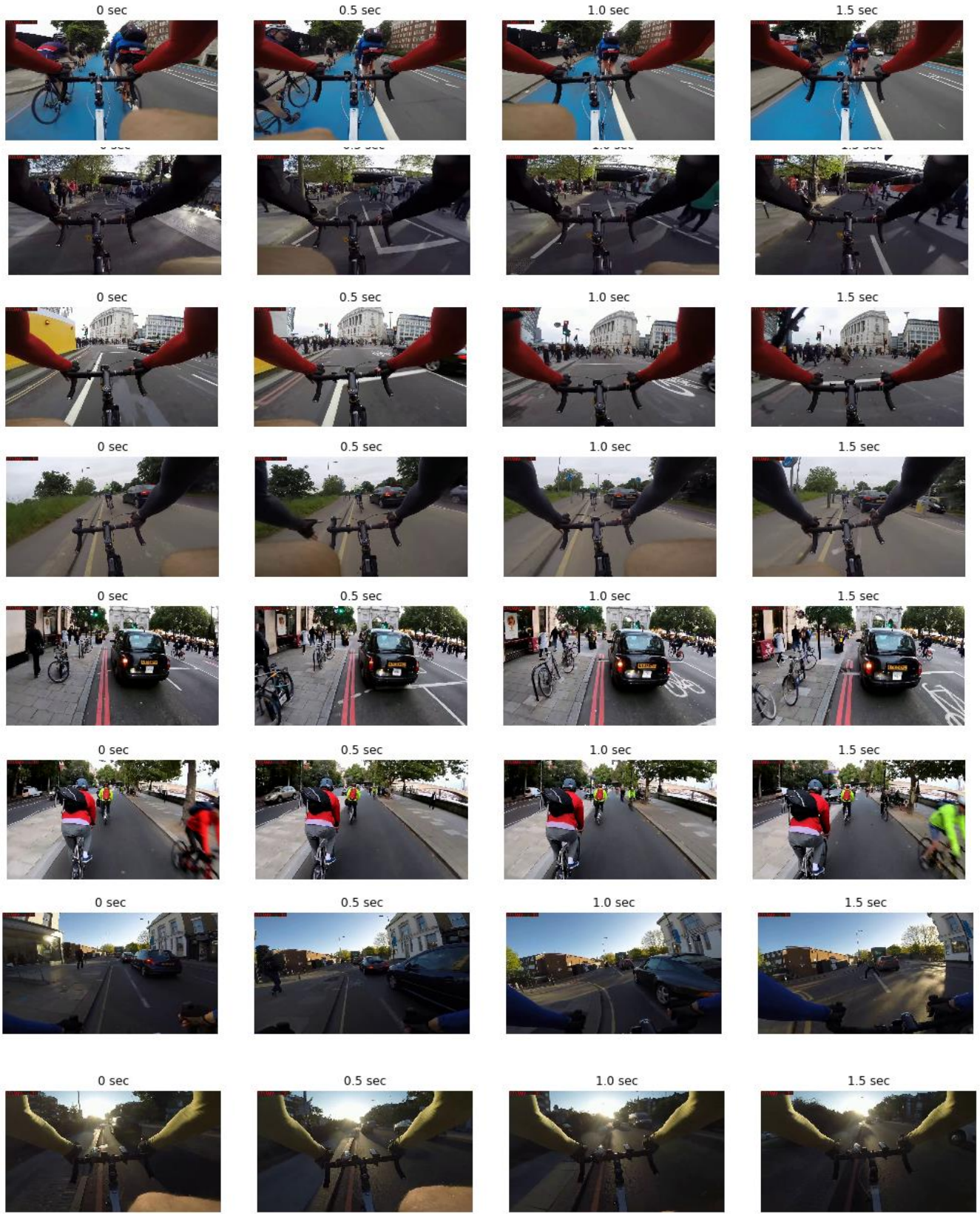

FIGURE 5
EXAMPLES OF PREDICTED CYCLING SAFE RIDE BY CYCLINGNET

## 5. DISCUSSION

*5.1 CyclingNet as the state-of-the-art method for detecting cycling near misses*

Understanding safety as a clue from the overall scene and interaction of different road-users remains a challenge. In this paper, we introduced CyclingNet as a novel method for detecting cycling near misses from video streams of moving bicycles in a complex urban setting. The model has shown strong performance in detecting near misses, regardless of the complexity of the scene, time of the day, weather, visual conditions, or the placement of the camera on the bike. Due to the absence of other models or benchmark datasets for the stated purpose, it remains a challenge to compare our results to other models, besides the ones we developed as base models. This, however, makes The CyclingNet model a vital and indispensable model for the field of road safety and more specifically, for detecting near misses. Accordingly, this makes it good practice for generalisation, deployment, and transfer learning to detect near misses for other road-users or other safety-related domains.

*5.2 Covid-19 pandemic and the increase in the number of people on bikes*

Whether temporarily or permanently, it has been debated that there is an increase in the numbers of people on bikes with different profiles and social characteristics (DfT, 2020).

There are no doubts that this increase in cycling would have direct benefits for health and the environment. With this increase, however, cycling infrastructure needs further preparation to host the increased numbers and more safety-related measures need to be considered. Accordingly, automating the detection of near misses could lead to drawing more significant safety policies for cycling in cities. Nevertheless, we need to understand the capacity of the current cycling infrastructure for a safe ride, and the tipping point for the increase in the number of near misses based on the interaction with other people on bikes or other road users.

*5.3 AI-embedded system for detecting near misses, their risk factors, and their causal inference*

Improving transport safety studies, especially for the most vulnerable road-users through automation is still a new domain of research. While detecting cycling near misses is vital for evaluating risky situations and better understanding the experiences of people on bikes, it is still one task towards understanding in depth the reasons for near misses, and their risk factors, and how to avoid them in future. According, AI-embedded system can be developed to capture not only near misses but also to understand the dynamics of the built and natural environments, in addition to detecting and localising either the various road users or the objects that could cause a risky situation for people on bikes. To do so, a pipeline of deep models can be utilities to sense, detect, and extract information of the different layers of the cities (the built environment, natural environment, transport, infrastructure) to draw significant conclusions (Ibrahim et al., 2020a). Different models have been proposed that could be utilised for the stated issue such as detecting the state of the environment and counting different road-users (Mohamed R Ibrahim et al., 2019) and recognising weather and visual conditions (Mohamed R. Ibrahim et al., 2019). After integrating these different deep models, a Bayesian approach for causal inference can be added to understand the effect and influence of each risk factor. We can also understand how such near miss experiences vary with the individual characteristics of the person who cycles such as their age, gender and even modifiable variables such as the level of training. In this respect, such models represent an efficient way to collect data. Accordingly, new transport safety policies can be taken into consideration, or a guideline for altering individual behaviours with riding bicycles or driving cars can be considered.

*5.4 Model limitations and future work*

The model shows a high validation for generalisation. However, there are still some limitations that need to be addressed in future work. First, instrumented bikes and groups of volunteered cyclists can be utilised for collecting new datasets which could offer a further verification of the introduced model. Second, introducing a new model to classify the different types of near misses after detection would allow a better understanding of the frequency of the different types of near misses. Third, developing the model to accurately extract the start and end of an event is another domain that needs further research. Lastly, applying similar models to detect safety measures and near misses for other road users such as pedestrians and car drivers would allow tailored-made policies or guidelines for the interaction of the different road-users according to the specific type of near misses for a given road-user.

## 6. REMARKS

The problem of detecting cycling near misses from video streams of moving bicycles in real-world settings has not been addressed in the current literature. In this paper, we utilised the advances of computer vision and deep learning to detect such events in a near real-time fashion. We introduced the CyclingNet model, a new deep computer vision for detecting cycling near misses from video streams of moving bicycles in complex urban environments. The model is structured as a single stream and trained in an end-to-end fashion, exploiting both single RGB frames, and optical flow data. After training the model of data of both of near misses and safe rides, the results show high performance on both training and validation data sets.

The model is intended to be used for generating information that can draw significant conclusions regarding cycling behaviour in cities and elsewhere, which could help planners and policy-makers to better understand the requirement of safety measures when designing infrastructure or drawing policies. As for future work, the model can be pipelined with other state-of-the-art classifiers and object detectors simultaneously to understand the causality of near misses based on factors related to interactions of road-users, the built and the natural environments.

7. ACKNOWLEDGEMENT

This work was supported by UCL Overseas Research Scholarship (ORS) and the Road Safety Trust (RST 38_03_2017). We would like to thank NVIDIA for the GPU grant.